\documentclass{article} 
\usepackage[preprint]{colm2026_conference}

\usepackage{microtype}
\usepackage{hyperref}
\usepackage{url}
\usepackage{graphicx}
\usepackage{booktabs}
\usepackage[textsize=small]{todonotes}
\usepackage{amsmath}
\usepackage{cleveref}
\usepackage{amsthm}
\usepackage{algorithm}      
\usepackage{algorithmicx}   
\usepackage{algpseudocode}
\usepackage{amssymb}
\usepackage{cleveref}
\usepackage{subcaption}

\usepackage{lineno}

\definecolor{darkblue}{rgb}{0, 0, 0.5}
\hypersetup{colorlinks=true, citecolor=darkblue, linkcolor=darkblue, urlcolor=darkblue}

\newtheorem{theorem}{Theorem}[section]

\theoremstyle{definition}
\newtheorem{definition}[theorem]{Definition}
\theoremstyle{remark}

\title{Skill-R1: Agent Skill Evolution via Reinforcement Learning}

\author{
Yash Vishe$^{1}$, Rohan Surana$^{1}$, Xunyi Jiang$^{1}$, Zihan Huang$^{1}$, Xintong Li$^{1}$, Nikki Lijing Kuang$^{1}$, \\
\textbf{Tong Yu$^{2}$, Ryan A. Rossi$^{2}$, Jingbo Shang$^{1}$, Julian McAuley$^{1}$, Junda Wu$^{1}$}
\\
$^{1}$UC San Diego \quad
$^{2}$Adobe Research \\
\texttt{\{yvishe,rsurana,xuj003,zih043,xil240,jshang,jmcauley,juw069\}@ucsd.edu} \\
\texttt{\{tyu,ryan.rossi\}@adobe.com} \\
}

\begin{document}

\ifcolmsubmission
\linenumbers
\fi

\maketitle

\begin{abstract}
Agentic large language models often rely on skills, reusable natural-language procedures that guide planning, action, and tool use.
In practice, however, skills are typically improved through prompt engineering or by aligning the task LLM itself to revised skills, 
which is costly, model-specific, and often infeasible for closed-source models.
Meanwhile, skill optimization is not a one-step prompt engineering problem, but a recurrent process with two coupled levels of credit assignment.
A useful skill must improve rollout quality under the current conditioning, while a useful revision must turn observed successes and failures into a better skill for the next round.
We therefore formulate skill evolution as a bi-level optimization problem over rollout selection within each generation and skill improvement across generations.
We propose \textbf{Skill-R1}, a reinforcement learning framework for instance-level recurrent skill optimization from verifiable rewards.
Rather than updating the task LLM, \textbf{Skill-R1} trains a lightweight skill generator that conditions on the task context, 
prior rollouts, and their verified outcomes to produce skills that steer a frozen task LLM.
This design preserves black-box compatibility with both open- and closed-source models while making adaptation substantially cheaper than model-level updates.
\textbf{Skill-R1} proceeds over multiple generations. At each generation, the current skill induces rollouts from the task LLM, whose verified outcomes are fed back to produce the next revision.
To optimize the skill generator over this recurrent process, we introduce a bi-level group-relative policy optimization objective with \textit{intra-generation} and \textit{inter-generation} advantages.
The intra-generation term compares rollouts under shared skill conditioning, while the inter-generation term rewards revisions that improve the induced behavior over successive generations.
Together, these terms provide a principled objective for directional skill evolution rather than one-shot or heuristic self-refinement.
Empirical evaluations suggest that Skill-R1 achieves consistent improvements over both no-skill baselines and standard GRPO across benchmarks with verifiable rewards. The gains are particularly strong on complex, multi-step tasks, where single-pass refinement methods struggle.
\end{abstract}

\section{Introduction}

Agentic large language models increasingly rely on external skills~\citep{xu2026agent,jiang2026sok,chen2026cua,nguyen2025guisurvey,huang2025towardsagentic}, 
which are reusable natural-language procedures that specify how an agent should decompose a task, invoke tools, and verify intermediate results. 
Since skills are designed as complementary modules of the task language model, 
they provide a practical interface for shaping LLM behavior without updating the task language model. 
This separation is practical in realistic deployments, where the task model may be proprietary~\citep{gao2025survey,huang2025surveyfoundation}, 
expensive to adapt~\citep{zhou2025memento,wu2024personalizedmllm,wu2025mitigatingforget,wang2025selfupdatable,zhang2024personalizationsurvey}, or shared across many downstream tasks~\citep{li2026organizing,chen2026skillcraft,jiang2026xskill}.

However, many skill-based systems rely on well-engineered skills as predefined and fixed artifacts, 
or revise them only through prompting and single-round self-refinement~\citep{jiang2026sok,xu2026agent,chen2026cua,jiang2026xskill,chen2026skillcraft,xia2026skillrl,zhang2026memskill}. 
These strategies can be misaligned and inefficient, since they rely on repeated prompt edits and are not directly aligned with downstream rewards~\citep{wu2025incontext,kveton2025activedpo,xia2025surveyactive}.
On the other hand, some recent works adapt the task language model to the skill by fine-tuning the task language model on the skill~\citep{wang2025reinforcement,li2025commit,wang2024instructgraph},
but this approach can be computationally costly and limited in adaptivity to many downstream tasks~\citep{zhou2025memento,xia2026skillrl,jiang2026adaptationagenticaisurvey,wu2025mitigatingforget,yao2024federatedllm}.
Moreover, skill optimization should not be a single prompt-editing step, but a recurrent decision problem: 
a useful skill should improve rollout quality under the current context, 
and a useful revision should transform observed successes and failures into a better skill for the next round. 
This naturally leads to a bi-level optimization objective that couples execution quality within each generation with skill improvement across generations.

To address this problem, we propose \textbf{Skill-R1}, a reinforcement learning framework for recurrent agent skill evolution from verifiable rewards. 
\textbf{Skill-R1} separates \emph{task execution} from \emph{skill improvement}. 
A frozen task LLM produces rollouts conditioned on the current skill, 
while a lightweight skill generator is trained to revise that skill based on the task context, prior rollouts, and their verified outcomes~\citep{xia2025knowledgequery,wang2025dice,vannguyen2025smallsurvey}. 
By updating only the skill generator, the framework remains compatible with both open- and closed-source task models and can be more efficient than updating the task model itself~\citep{vannguyen2025smallsurvey,xia2025surveyactive}.

The core interaction loop of \textbf{Skill-R1} is multi-generation. 
For a given instance, the current skill induces a group of rollouts from the frozen task model, a verifier scores these rollouts, and the resulting trials and errors~\citep{wu2025ocean, wu2024decot},
and reward signals are appended to an instance-specific history that conditions the next skill revision. 
Recurring execution of this process yields an evolutionary view of skill improvement, in which each generation proposes a new skill, 
tests it through a rollout population, and passes forward evidence for subsequent refinement.
This recurrent structure makes it possible to optimize directional improvement~\citep{xia2026skillrl,sun2025seagent,zhou2025memento} on the same instance rather than selecting among isolated one-shot attempts.

The policy optimization of the skill generator requires both intra-generation and inter-generation credit assignment to optimize the skill generator.
To enable optimization of the recurrent skill generator, we introduce a bi-level group-relative policy optimization~\citep{zhong2026rc,shao2024deepseekmath,mundada2026wsgrpo} objective that mirrors the recurrent structure above. 
The \emph{intra-generation} term compares rollouts sampled under the same skill and captures relative execution quality within a generation. 
The \emph{inter-generation} term measures whether a revised skill improves the average verified performance relative to the previous generation.
Together, these signals provide a principled objective for rewarding both strong rollouts and beneficial skill revisions~\citep{wu2025incontext, surana2026massdpo, li2025impsamplingdpo, huang2025pluralistic}.
Our experiments demonstrate that Skill-R1 consistently improves agent performance across diverse reasoning and tool-use benchmarks, outperforming both no-skill baselines and standard GRPO. 
We also show a clear performance gap between skill generations, which demonstrates the effectiveness of the recurrent skill evolution.
We summarize the contributions of this work as follows:
\begin{itemize}
    \item We formulate recurrent instance-level skill evolution as a bi-level group-relative policy optimization problem.
    \item We propose \textbf{Skill-R1}, a model-agnostic framework that trains a lightweight skill generator while keeping the task LLM frozen.
    \item We introduce a recurrent multi-generation rollout process together with a bi-level GRPO objective that combines \emph{intra-} and \emph{inter-generation} advantages. 
    \item We evaluate \textbf{Skill-R1} on agent tasks with verifiable rewards, comparing with agent skill and GRPO baselines and achieving superior performance.
\end{itemize}
\section{Preliminaries}\label{sec:preliminaries}

\subsection{Agent Skills} \label{sec:prelim-skills}
Recent LLM-agent systems commonly treat skills as reusable units of external procedural knowledge that can be retrieved, executed, and updated without modifying the underlying task model. 
Following this practice, let $\pi_{\mathrm{task}}$ denote a frozen task LLM, and let $\mathcal{S}$ be a space of explicit skill artifacts.

\begin{definition}[Agent Skill]
An agent skill is a reusable procedural artifact $s \in \mathcal{S}$ that conditions the execution of the task model on an input instance. 
A skill may encode natural-language instructions, structured workflows, tool-use guidance, or other procedural constraints, while remaining external to the parameters of $\pi_{\mathrm{task}}$.
\end{definition}

Given an instance $x$ and a skill $s$, the frozen task model induces a rollout distribution
\begin{equation} \label{eq:rollout_dist}
\pi_{\mathrm{task}}(\tau \mid x, s),
\end{equation}
where $\tau$ denotes the resulting trajectory. 
In our setting, skills evolve recurrently on the same instance. 
We write $s_g \in \mathcal{S}$ for the skill used at generation $g$, and $\mathcal{H}_g$ for the instance-specific history available up to generation $g$, such as prior rollouts and their verified outcomes. 
These variables will be used to formalize skill evolution in later sections.

\subsection{Group-Relative Policy Optimization} \label{sec:prelim-grpo}
We briefly review Group-Relative Policy Optimization (GRPO) in the language-generation setting, where a policy $\pi_\theta(\cdot \mid x)$ defines a distribution over response rollouts $y$ conditioned on a prompt or context $x \sim \mathcal{D}$. 
GRPO samples a group
\begin{equation} \label{eq:group_sampling}
\mathcal{G}(x)=\{y^{(i)}\}_{i=1}^K,
\qquad
y^{(i)} \stackrel{\mathrm{i.i.d.}}{\sim} \pi_\theta(\cdot \mid x),
\end{equation}
assigns each rollout a verifier score $r(y^{(i)})$, and defines the group-relative advantage
\begin{equation} \label{eq:grpo_adv}
A\!\left(y^{(i)};\mathcal{G}(x)\right)
=
r\!\left(y^{(i)}\right)
-
\frac{1}{K}\sum_{j=1}^K r\!\left(y^{(j)}\right).
\end{equation}
The GRPO objective is
\begin{equation} \label{eq:grpo_obj}
\mathcal{L}_{\mathrm{GRPO}}(\theta)
=
\mathbb{E}_{x \sim \mathcal{D},\, \mathcal{G}(x)}
\left[
\sum_{i=1}^K
A\!\left(y^{(i)};\mathcal{G}(x)\right)
\log \pi_\theta\!\left(y^{(i)} \mid x\right)
\right].
\end{equation}
This group-centered objective promotes rollouts that outperform their peers under the same input context.

\section{Skill-R1: Agent Skill Evolution via Reinforcement Learning}
\label{sec:methodology}

\begin{figure}[ht]
\centering 
\includegraphics[width=1\linewidth]{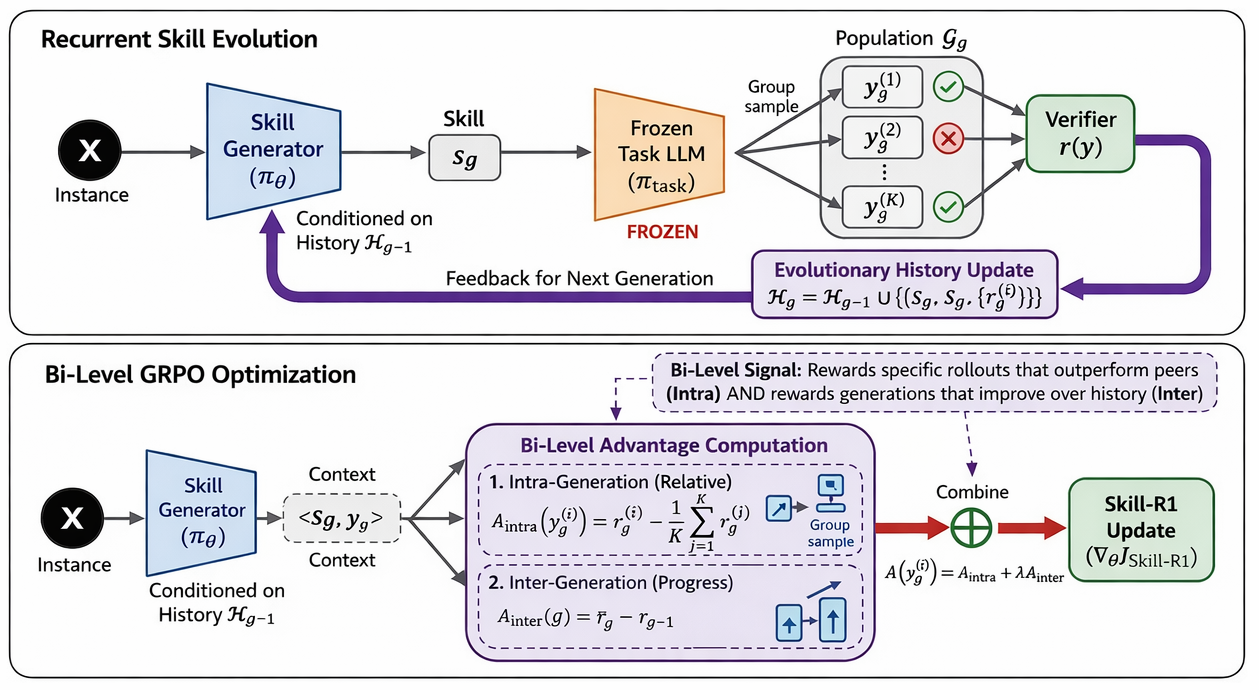} 
\caption{Overview of Skill-R1. \textbf{Recurrent Skill Evolution} iteratively generates skills, rolls out a frozen task LLM, and stores verifier-scored outcomes in an evolutionary history. \textbf{Bi-level GRPO Optimization} computes GRPO advantages from both intra-generation relative performance and inter-generation progress, and uses it to update the skill generator while keeping the task LLM frozen.}
\label{fig:flow} 
\end{figure}

We propose \textbf{Skill-R1} (illustrated in~\Cref{fig:flow}), a reinforcement learning framework that improves a frozen task model by recurrently refining external skills without updating the task model itself.
\textbf{Skill-R1} decouples \emph{task execution} from \emph{skill improvement}: a frozen task LLM generates rollout trajectories conditioned on a selected skill,
while a separate editor policy updates skills based on verified execution outcomes.

    
\subsection{Problem Setup} \label{sec:method-setup}

Let $x \in \mathcal{X}$ denote a task instance, and let $\mathcal{B}_x \subseteq \mathcal{S}$ denote the skill bank for instance $x$. 
Following \Cref{sec:prelim-skills}, for a selected skill $s_0 \in \mathcal{B}_{x}$, 
the frozen task model $\pi_{\mathrm{task}}$ induces a rollout distribution \Cref{eq:rollout_dist} over trajectories $y^{(i)} \in \mathcal{G}(x,s_0)$ group sampled (in \Cref{eq:group_sampling}) under skill conditioning, 
while a verifier $f_X:\mathcal{G}(x,s_0)\to\mathbb{R}$ assigns each rollout a scalar reward $r^{(i)}$. 
We also maintain the generation-wise history
\begin{equation} \label{eq:history_init}
\mathcal{H}_0 = \{(x, s_0,y^{(i)},r^{(i)})\}_{i=1}^{K},
\qquad 
y^{(i)} \sim \pi_{\mathrm{task}}(\cdot \mid x, s_0), 
\qquad
i=1,\dots,K.
\end{equation}
Thus, the instance-level performance induced by $s_0$ is determined by the rollout policy and verifier jointly,
and our goal is to improve performance on the given instance by revising skills so that later rollout populations achieve higher verifier reward.

For each generation $g \ge 1$, a learnable skill generator $\pi_\theta$ produces the next skill conditioned on the current instance and the accumulated history:
\begin{equation} \label{eq:skill_gen}
\tau = \{(s_g,\mathcal{G}_g(x,s_g),\{r_g^{(i)}\}_{i=1}^{K})\}_{g=1}^{G},
\qquad
s_g \sim \pi_\theta(\cdot \mid x, \mathcal{H}_{g-1}),
\qquad g=1,\dots,G.
\end{equation}
This defines a recurrent skill-evolution process in which each generated skill is evaluated through the rollout population it induces under the frozen task model \Cref{eq:rollout_dist}, 
and the resulting verifier outcomes are appended to the history to guide subsequent revisions. 
Then the recurrent skill evolution problem can be defined as optimizing $\pi_\theta$ to maximize verifier reward across the entire recurrent process:
\begin{equation} \label{eq:rl_objective}
J(\theta)
=
\mathbb{E}_{x \sim p(x)}
\mathbb{E}_{\substack{
s_g \sim \pi_\theta(\cdot \mid x, \mathcal{H}_{g-1}),\:
y_g^{(i)} \sim \pi_{\mathrm{task}}(\cdot \mid x, s_g)
}}
\left[
\sum_{g=1}^{G}
\gamma^{g-1}
\frac{1}{K}
\sum_{i=1}^{K} r_g^{(i)}
\right].
\end{equation}
Directly optimizing the objective in \eqref{eq:rl_objective} is challenging because the reward signal is evaluated on the task model's rollouts rather than directly on the generated skill. 
To address this structural decoupling between skill generation and task execution, we extend the standard GRPO framework to a bi-level setting. 
In the following section, we formulate a bi-level GRPO objective and derive \emph{bi-level advantages}. 
This formulation allows us to properly assign credit to the skill generator $\pi_\theta$ based on the relative, aggregate performance of the rollout populations that each skill induces.

\begin{algorithm}[ht]
\caption{Skill-R1: Agent Skill Evolution via Reinforcement Learning}
\label{alg:skillr1}
\begin{algorithmic}[1]
\Require Task distribution $p(x)$, frozen task LLM $\pi_{\mathrm{task}}$, skill generator $\pi_\theta$, initial skill bank $\mathcal{B}$, verifier $f$, generations $G$, group size $K$, mixing coefficient $\lambda$
\For{each task instance $x \sim p(x)$ and its skill bank $\mathcal{B}_x$}
    \State Select initial skill $s_0 \in \mathcal{B}_x$
    \State Sample rollouts $\mathcal{G}_0(x, s_0) = \{y_0^{(i)}\}_{i=1}^K$ from $\pi_{\mathrm{task}}(\cdot \mid x, s_0)$
    \State Score $r_0^{(i)} = f(y_0^{(i)})$ for each $i$; initialize $\mathcal{H}_0 = \{(x, s_0, y_0^{(i)}, r_0^{(i)})\}_{i=1}^K$
    \For{$g = 1, \dots, G$}
        \State Generate skill $s_g \sim \pi_\theta(\cdot \mid x, \mathcal{H}_{g-1})$ \Comment{\Cref{eq:skill_gen}}
        \State Sample rollouts $\mathcal{G}_g(x, s_g) = \{y_g^{(i)}\}_{i=1}^K$ from $\pi_{\mathrm{task}}(\cdot \mid x, s_g)$
        \State Verify $r_g^{(i)} = f(y_g^{(i)})$ for each $i$
        \State Compute $A_{\mathrm{intra}}(y_g^{(i)})$, $A_{\mathrm{inter}}(g)$, and $A(y_g^{(i)})$ \Comment{\Cref{eq:intra_adv,eq:inter_adv,eq:bilevel_adv}}
        \State Update $\mathcal{H}_g = \mathcal{H}_{g-1} \cup \{(s_g, \mathcal{G}_g, \{r_g^{(i)}\}_{i=1}^K)\}$
    \EndFor
\EndFor
\State Update $\theta$ by maximizing $\mathcal{L}_{\mathrm{Skill\text{-}R1}}(\theta)$ over accumulated rollouts \Comment{\Cref{eq:skillr1_obj}}
\end{algorithmic}
\end{algorithm}

\subsection{Bi-Level Group-relative Advantages }
\label{sec:method-advantage}

The reward signal in \Cref{eq:rl_objective} is evaluated on rollouts from the frozen task model, yet the trainable parameters reside in the skill generator $\pi_\theta$.
To route rollout-level feedback to the skill that induced it, we decompose credit assignment into two complementary advantage signals.
Within generation $g$, all $K$ rollouts share the same skill $s_g$, so we apply the group-relative comparison from \Cref{eq:grpo_adv} directly:
\begin{equation} \label{eq:intra_adv}
A_{\mathrm{intra}}(y_g^{(i)})
=
r_g^{(i)} - \frac{1}{K}\sum_{j=1}^K r_g^{(j)}.
\end{equation}
Centering by the group mean makes this signal invariant to absolute reward scale and isolates rollout quality from skill quality.
However, $A_{\mathrm{intra}}$ is blind to whether a skill revision improved on its predecessor.
To capture cross-generation progress, we define the population mean reward and the inter-generation advantage
\begin{equation} \label{eq:inter_adv}
A_{\mathrm{inter}}(g) = \bar{r}_g - \bar{r}_{g-1}, \qquad \bar{r}_g = \frac{1}{K}\sum_{i=1}^K r_g^{(i)}
\end{equation}
with $A_{\mathrm{inter}}(1)=0$.
A positive value indicates that the revised skill shifted the rollout distribution toward higher reward; a negative value penalizes regressions.
We combine the two signals into a single bi-level advantage:
\begin{equation}
A(y_g^{(i)}) = A_{\mathrm{intra}}(y_g^{(i)}) + \lambda\, A_{\mathrm{inter}}(g),
\label{eq:bilevel_adv}
\end{equation}
where $\lambda \ge 0$ interpolates between pure within-generation GRPO ($\lambda{=}0$) and a regime that additionally rewards cross-generation improvement.

\subsection{GRPO Objective for Skill Evolution}
\label{sec:method-objective}

We optimize $\pi_\theta$ with a clipped GRPO surrogate that aggregates the bi-level advantage across all generations.
Let $\rho_g = \pi_\theta(s_g \mid x, \mathcal{H}_{g-1}) / \pi_{\theta_{\mathrm{old}}}(s_g \mid x, \mathcal{H}_{g-1})$ be the importance ratio between the current and data-collection policies. The \textbf{Skill-R1} objective is
\begin{equation}
\begin{aligned}
\mathcal{L}_{\mathrm{Skill\text{-}R1}}(\theta)
&=
\mathbb{E}_{x \sim p(x)}
\Bigg[
\sum_{g=1}^{G}
\sum_{i=1}^{K}
\min\!\left(
  \rho_g\, A(y_g^{(i)}),\;
  \mathrm{clip}(\rho_g,\,1{-}\epsilon,\,1{+}\epsilon)\, A(y_g^{(i)})
\right)
\\
&\quad
-
\beta\,\mathrm{KL}\!\left(\pi_\theta(\cdot \mid x, \mathcal{H}_{g-1}) \,\|\, \pi_{\mathrm{ref}}(\cdot \mid x, \mathcal{H}_{g-1})\right)
\Bigg],
\end{aligned}
\label{eq:skillr1_obj}
\end{equation}
where $A(y_g^{(i)})$ is the bi-level advantage from \Cref{eq:bilevel_adv}, $\epsilon > 0$ is the clipping radius, and $\beta \ge 0$ weights a KL penalty anchoring the policy near a reference $\pi_{\mathrm{ref}}$.
Summing over $g$ generations couples the objective across the entire recurrent process in \Cref{eq:skill_gen},
and thus $\pi_\theta$ is trained to produce improving skill trajectories rather than isolated single-step revisions.
Because only $\pi_\theta$ is updated, the task LLM $\pi_{\mathrm{task}}$ remains frozen and can be open- or closed-source and no gradients through $\pi_{\mathrm{task}}$ are required.
The full procedure is summarized in \Cref{alg:skillr1}.

\section{Experiments}
\label{sec:experiments}

Our experiments are designed to address three key questions: (1) whether conditioning a frozen task language model on progressively evolving skills yields measurable gains over a no-skill baseline, (2) whether a multi-generation rollout framework provides benefits beyond standard GRPO training~\citep{shao2024deepseekmath}, and (3) whether a trained skill editor outperforms an inference-only editor without gradient-based optimization. 
\subsection{Experimental Setup}
\paragraph{Tasks.}We evaluate Skill-R1 on two benchmarks requiring multi-step reasoning, tool use, and verifiable answers: GAIA~\citep{mialon2024gaia}, which consists of 165 real-world tasks over heterogeneous sources (e.g., PDFs, spreadsheets, and web pages), and WebWalker~\citep{wu2025webwalker}, which focuses on multi-hop web navigation and cross-page reasoning. Together, they capture both structured reasoning and interactive decision-making.

\paragraph{Baselines.} Our baselines are designed to isolate key components of the pipeline. No Skills measures the base model’s standalone capability without skill conditioning. Vanilla GRPO applies standard Group Relative Policy Optimization to train the editor without bi-level advantage decomposition or multi-generation rollouts, evaluating whether conventional RL alone suffices to improve skill quality.

\paragraph{Skill initialization.} 
Base skills are constructed via a two-stage distillation procedure applied per benchmark. First, GPT-4o-mini is run on a seed set of ten tasks to collect reasoning traces, including both successful and failed trajectories. Second, these traces are provided to a stronger LLM (Claude Opus 4.6), which abstracts recurring patterns into concise, reusable skill guides. For reproducibility, each Skill-R1 run uses an isolated copy of the skill directory, while baselines share a read-only version.
\paragraph{Training vs. Inference Decomposition.} Skill-R1 (Inference) runs the full multi-generation rollout pipeline but uses a frozen Qwen editor, thereby isolating the effect of the rollout framework from any gradient-based learning. Skill-R1 (GRPO) represents the full system, where the Qwen-based editor is trained online with GRPO and bi-level advantages, capturing the additional gains from learned skill refinement.

All experiments use GPT-4o-mini as the task-solving model. 

\subsection{Results for GAIA}

\begin{table}[ht]
\centering
\resizebox{\textwidth}{!}{%
\begin{tabular}{lcccc}
\toprule
\textbf{Method} & \textbf{Level 1} ($n\!=\!53$) & \textbf{Level 2} ($n\!=\!86$) & \textbf{Level 3} ($n\!=\!26$) & \textbf{Overall} ($n\!=\!165$) \\
\midrule
GPT-4o-mini (no skills)             & 4/53 \;\;(7.5\%)  & 6/86 \;\;(7.0\%)  & 0/26 \;\;(0.0\%)  & 10/165 \;\;(6.1\%)  \\
Vanilla GRPO (Qwen3-4b)                    & 21/53 \;\;(39.6\%)   & 24/86 \;\;(27.9\%)   & 4/26 \;\;(15.4\%)   & 49/165 \;\;(29.7\%)   \\

Skill-R1 (Inference, Qwen3-4b)      & 22/53 (41.5\%)     & 21/86 (24.4\%)     & 8/26 \;\;(30.8\%)  & 51/165 (30.9\%)     \\
Skill-R1 (GRPO, Qwen3-4b)           & \textbf{31/53 (58.5\%)} & \textbf{28/86 (32.6\%)} & \textbf{10/26 (38.5\%)} & \textbf{69/165 (41.8\%)} \\
\bottomrule
\end{tabular}%
}
\caption{Main results on the GAIA benchmark (165 tasks).}
\label{tab:main_results}
\end{table}

Tables~\ref{tab:main_results} presents the results on GAIA. GPT-4o-mini (no skills) performs poorly, achieving only $6.1\%$ accuracy, highlighting its difficulty with multi-step reasoning over heterogeneous sources. Vanilla GRPO substantially improves performance to $29.7\%$, with gains across all difficulty levels, but remains limited on more complex tasks, particularly Level~3 ($15.4\%$), indicating insufficient ability to compose and reuse structured knowledge. Skill-based reasoning further improves performance. The inference-only Skill-R1 setup achieves $30.9\%$, suggesting that the multi-generation rollout framework itself provides benefits beyond standard RL. Training the editor yields a larger improvement to $41.8\%$, a $+12.1$ point gain over Vanilla GRPO. The improvements are most prominent on harder tasks. Level~3 accuracy increases to $38.5\%$, compared to $0.0\%$ for the no-skill baseline and $15.4\%$ for Vanilla GRPO. This highlights the importance of learned skill refinement in enabling compositional reasoning over complex problems.

\subsection{Results for WebWalker}

\begin{table}[ht]
\centering
\resizebox{\textwidth}{!}{%
\begin{tabular}{lcccc}
\toprule
\textbf{Method} & \textbf{Easy} ($n\!=\!8$) & \textbf{Medium} ($n\!=\!68$) & \textbf{Hard} ($n\!=\!24$) & \textbf{Overall} \\
\midrule
GPT-4o-mini (no skills)           & 0/8 \;\;(0.0\%)           & 1/68 \;\;(1.5\%)        & 1/24 \;\;(4.2\%)        & 2/100 \;\;(2.0\%) \\
Vanilla GRPO (Qwen3-4b)          & 1/8 \;\;(12.5\%)          & 15/68 (22.1\%)          & 6/24 (25.0\%)          & 22/100 (22.0\%) \\
Skill-R1 (Inference, Qwen3-4B)   & 0/8 \;\;(0.0\%)           & 14/68 (20.6\%)          & 5/24 (20.8\%)          & 19/100 (19.0\%) \\
Skill-R1 (GRPO, Qwen3-4B)        & \textbf{1/8 \;\;(12.5\%)} & \textbf{20/68 (29.4\%)} & 5/24 (20.8\%)          & \textbf{26/100 (26.0\%)} \\
\bottomrule
\end{tabular}%
}
\caption{Results on the WebWalker benchmark (100 tasks) by difficulty.}
\label{tab:webwalker_results}
\end{table}
\begin{table}[ht]
\centering
\resizebox{\textwidth}{!}{%
\begin{tabular}{lccc}
\toprule
\textbf{Method} & \textbf{Multi-source} ($n\!=\!63$) & \textbf{Single-source} ($n\!=\!37$) & \textbf{Overall} \\
\midrule
GPT-4o-mini (no skills)           & 1/63 \;\;(1.6\%)         & 1/37 \;\;(2.7\%)         & 2/100 \;\;(2.0\%) \\
Vanilla GRPO (Qwen3-4b)          & 14/63 (22.2\%)           & 8/37 \;\;(21.6\%)        & 22/100 (22.0\%) \\
Skill-R1 (Inference, Qwen3-4B)   & 11/63 (17.5\%)           & 8/37 \;\;(21.6\%)        & 19/100 (19.0\%) \\
Skill-R1 (GRPO, Qwen3-4B)        & \textbf{18/63 (28.6\%)}  & 8/37 \;\;(21.6\%)        & \textbf{26/100 (26.0\%)} \\
\bottomrule
\end{tabular}%
}
\caption{Results on the WebWalker benchmark (100 tasks) by question type.}
\label{tab:webwalker_source}
\end{table}

Tables~\ref{tab:webwalker_results} and~\ref{tab:webwalker_source} present the results on WebWalker. GPT-4o-mini (no skills) achieves only $2.0\%$ accuracy, reflecting the difficulty of multi-hop web navigation and cross-page reasoning. Vanilla GRPO improves performance to $22.0\%$, with consistent gains across difficulty levels. However, performance remains constrained in settings requiring deeper information synthesis, particularly in medium and multi-source tasks. Skill-based reasoning further enhances performance. The inference-only Skill-R1 setup achieves $19.0\%$, showing competitive performance without any gradient-based learning, while the GRPO-trained editor improves results to $26.0\%$, yielding a $+4.0$ percentage point gain over Vanilla GRPO. Improvements are especially evident in medium ($29.4\%$) and multi-source ($28.6\%$) settings, indicating that learned skill refinement enables more effective navigation and aggregation of information across multiple pages. An interesting observation is the relatively lower accuracy on the easy subset. This is primarily due to its small size and sensitivity to brittle navigation failures, such as incomplete page retrieval. As a result, minor errors disproportionately affect performance, leading to higher variance compared to other difficulty levels.

\section{Analysis}
\label{sec:analysis}

\subsection{Reward and Accuracy Progression Across Generations}

\begin{figure}[ht]
    \centering
    \includegraphics[width=\linewidth]{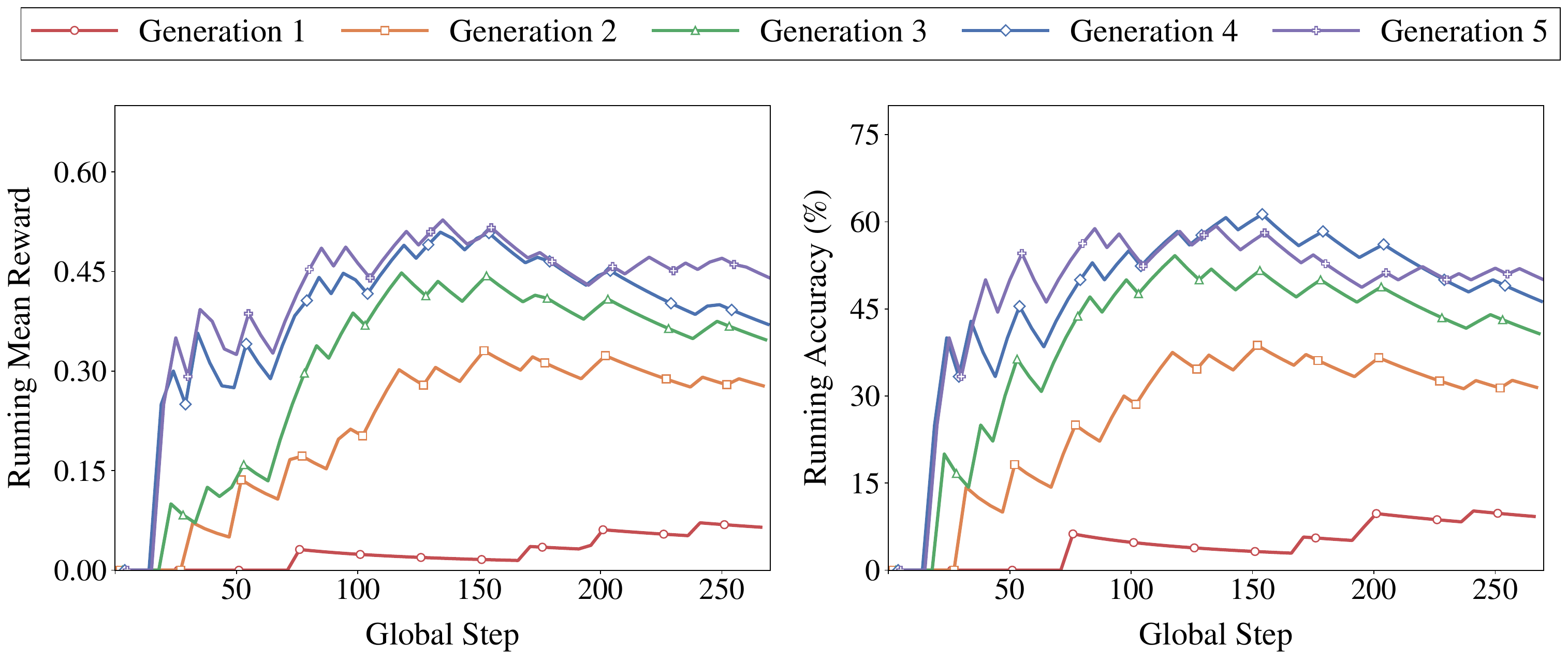}
    \caption{Accuracy and reward curves across 5 generations. }
    \label{fig:combined}
    \vspace{-1em}
\end{figure}

To better understand the dynamics of skill evolution, we analyze both the per-task running mean reward and running accuracy across generations (\Cref{fig:combined}). The reward curve reflects the average verifier score over rollout populations, while the accuracy curve indicates whether at least one rollout successfully solves a task. Although rewards are binary in our setting, the two metrics capture different aspects of performance. Reward reflects the consistency of successful rollouts within a task, whereas accuracy measures only whether a task is solved at least once. As a result, the two trends are correlated but not identical, providing complementary views of both capability and reliability.

The two figures reveal a clear and largely monotonic ordering after the initial warm-up: Generation~5 outperforms Generation~4, which outperforms Generation~3, and so on, but the magnitude of improvement is not uniform across generations. The largest gains occur between Generations~1 and~3, where both reward and accuracy increase sharply as useful skills begin to transfer across tasks. By the end, Generation~5 reaches a running mean reward of $0.44$ and accuracy of $50.0\%$, compared to $0.06$ reward and $9.3\%$ accuracy for Generation~1, while Generation~4 achieves $0.37$ and $46.3\%$. This pattern suggests that most of the broad performance gains are realized by the first three generations, while Generations~4 and~5 increasingly plateau and primarily refine an already strong skill set rather than opening up a comparably new level of capability.

Comparing reward and accuracy further clarifies this behavior. The larger improvement in reward than accuracy from Generation~4 to~5 indicates that later generations improve consistency rather than merely solving additional tasks. Once a task becomes solvable, they increase the likelihood that sampled rollouts succeed, enhancing reliability within each task. The largest jump still occurs between the early generations, especially from Generation~1 to Generations~2 and~3, indicating that the highest-value skill repairs are discovered early. Later generations improve more modestly, but the accuracy panel shows that those improvements remain meaningful rather than cosmetic. This suggests that a small number of editing rounds captures the majority of improvements, with additional iterations yielding incremental but meaningful gains in robustness and coverage.

\subsection{Qualitative Analysis}

\subsubsection{Trained v/s Inference Skill Editor}

The performance gap between Skill-R1 (GRPO) and Skill-R1 (Inference) in \Cref{tab:main_results,tab:webwalker_results} reflects a fundamental difference in how each editor interprets rollout evidence. The trained editor treats rollouts as signals for targeted repair, while the inference-only editor tends to overfit to incidental context.

Using the \texttt{pdb} skill as a case study, the trained editor preserves the core executable structure and introduces focused safeguards around genuine failure modes (e.g., missing files or dependencies). In contrast, the inference-only editor rewrites the skill around spurious details from sampled rollouts, often discarding essential components such as executable code. This leads to brittle, task-specific artifacts that fail to generalize.

This qualitative difference is reflected in performance, while inference-only Skill-R1 provides modest gains over Vanilla GRPO on GAIA (30.9\% vs. 29.7\%), it underperforms on WebWalker (19.0\% vs. 22.0\%), the GRPO-trained editor consistently improves performance across benchmarks (41.8\% on GAIA and 26.0\% on WebWalker), demonstrating that reward driven editing better captures transferable structure.

Overall, training enables selective retention of useful behaviors while suppressing noise, leading to more robust and generalizable skill updates. Full skill listings for both the GRPO-edited and inference-edited \texttt{pdb} skill (alongside the original) are provided in \Cref{app:pdb_skill_comparison}.

\subsubsection{Editing Trajectory Across Generations}

Skill evolution follows a structured progression rather than simple accumulation. Early generations yield the largest gains by correcting high-impact errors, as also reflected in the sharp improvements from Generation 1 to 3 in \Cref{fig:combined}. Later generations provide smaller but consistent improvements by refining reliability and stabilizing behavior.

Qualitatively, this corresponds to a prune then specialize dynamic. Initial skills, which resemble broad reference documents, are quickly compressed into shorter executable routines. Subsequent generations focus on recurring failure modes and introduce tighter constraints and guardrails, resulting in compact, deployment oriented policies.

This pattern aligns with the quantitative trends: most performance gains are realized early, while later iterations primarily improve consistency rather than expanding capability. Together, these observations suggest that multi-generation evolution is effective not only for discovering better skills, but also for making them more reliable under repeated execution.
\section{Related Work}

\subsection{Agent Skill in Language Models}


Recent work treats agent skills as reusable procedural abstractions that guide LLM behavior without modifying model parameters. Surveys (~\citep{xu2026agent},~\citep{jiang2026sok}) characterize skills as structured intermediates between tool usage and agent coordination. On the acquisition side, prior work has focused on constructing and organizing skill libraries~\citep{nguyen2025guisurvey, huang2025surveyfoundation}. SkillWeaver(~\citep{zheng2025skillweaver}) enables agents to autonomously discover and distill skills through web interaction, while ASI(~\citep{wang2025inducing}) shows that programmatic skill representations improve reliability via verifiability. Similarly, CUA-Skill~\citep{chen2026cua} introduces a repository of parameterized execution graphs for computer-use agents. Adjacent agentic-LLM work treats reusable procedures and contextual conditioning as central design tools for tool use, document reasoning, and multimodal interaction~\citep{xia2025sand, wu2025docreact,wu2025ctrls,wang2025dice,huang2025towardsagentic}. Other works study orchestration and system-level design ~\citep{li2026organizing}, which explores ecosystem-scale skill coordination. More recent efforts move beyond static skill banks toward skill evolution. MemSkill~\citep{zhang2026memskill} models memory operations as skills refined through iterative control loops, while Memento-Skills~\citep{zhou2025memento} enables continual skill improvement through reflective updates without modifying the base model. However, these approaches largely rely on heuristic refinement or implicit feedback signals.

\subsection{Reinforcement Learning for Language Models}

Reinforcement learning has become a central paradigm for aligning language models with task objectives and feedback signals. Methods such as PPO ~\citep{schulman2017proximal} and its variants optimize policies using scalar rewards, while more recent approaches like GRPO ~\citep{shao2024deepseekmath} introduce group-relative comparisons to stabilize training and improve sample efficiency. Recent variants further extend GRPO to weakly-supervised and reward-conditioned settings~\citep{mundada2026wsgrpo,zhong2026rc}. Several works extend RL to agent settings and skill learning. SkillRL~\citep{xia2026skillrl} jointly learns hierarchical skills and policies through recursive reinforcement learning, enabling agents to acquire reusable behaviors over time. Despite these advances, most existing methods focus on optimizing the task model itself or operate within a single-generation setting, limiting their ability to capture iterative improvement.

\section{Conclusion}
\label{sec:conclusion}
We presented Skill-R1, a framework for recurrent skill evolution that improves agent performance without modifying the underlying task model. By decoupling execution from skill refinement and introducing a bi-level optimization objective, Skill-R1 enables iterative, reward-driven improvement over multiple generations. Empirically, Skill-R1 consistently outperforms both no-skill baseline and standard GRPO across benchmarks, with the largest gains observed on complex, multi-step tasks. Analysis shows that early generations drive most capability improvements, while later iterations enhance reliability and consistency, highlighting the importance of both evolution and stabilization. Overall, our results suggest that skill evolution is a practical alternative to model-level adaptation, offering a lightweight yet effective mechanism for improving agent behavior in both open- and closed-source settings.
\textbf{LLM usage disclosure:} AI tools were used to assist creating illustrative figures. 
\footnote{Tong and Ryan contributed to the conceptual and methodological design and did not process, store, or direct the use of project models or data.}

\bibliography{colm2026_conference}
\bibliographystyle{colm2026_conference}

\appendix

\section{Appendix}

\subsection{PDB Skill}
\label{app:pdb_skill_comparison}

This appendix provides the complete content of the \texttt{pdb} skill at three points in time, as referenced in \Cref{sec:analysis}: (1) the original skill from the shared skill library, (2) the version produced by the GRPO-trained Qwen3-4B editor after three generations of evolution (run \texttt{20260320\_131744\_0.25}), and (3) the version produced by the untrained base Qwen3-4B model operating in inference mode (run \texttt{20260321\_185812}).

\subsubsection{Original \texttt{pdb} Skill}

The original skill is a general-purpose reference covering BioPython's full API. We reproduce the header description and the Quick Start section to illustrate the skill's initial breadth; the remainder of the document (Dependencies, PDB File Format Basics, Python Libraries, Manual Parsing, Common Tasks, Quick Reference table) is analogous in scope.

\begin{small}
\begin{verbatim}
name: pdb
description: Use this skill whenever the user wants to read, parse, analyze,
  or manipulate PDB (Protein Data Bank) files. This includes extracting atomic
  coordinates, analyzing protein structures, identifying residues, chains,
  ligands, computing distances between atoms, visualizing molecular structures,
  and converting between molecular file formats. If the user mentions a .pdb
  file or asks about protein/molecular structure data, use this skill.

# PDB (Protein Data Bank) File Processing Guide

## Quick Start

```python
from Bio.PDB import PDBParser

parser = PDBParser(QUIET=True)
structure = parser.get_structure("protein", "structure.pdb")

# Iterate through the hierarchy: Structure -> Model -> Chain -> Residue -> Atom
for model in structure:
    for chain in model:
        print(f"Chain {chain.id}: {len(list(chain.get_residues()))} residues")
```

## PDB File Format Basics

PDB files are fixed-width text files. Key record types:
| ATOM   | Standard residue atom coordinates  |
| HETATM | Non-standard residue (ligands, water) |
| SEQRES | Full sequence of residues          |
| REMARK | Comments and metadata              |

[... BioPython parsing examples, distance computation, ligand extraction,
     sequence extraction, manual ATOM-record parsing, and quick-reference
     table omitted for brevity -- all preserved in the original file ...]
\end{verbatim}
\end{small}

\subsubsection{GRPO-Trained Editor: Edited \texttt{pdb} Skill}

\begin{small}
\begin{verbatim}
name: pdb
description: [unchanged from original]

# PDB (Protein Data Bank) File Processing Guide

## Quick Start

```python
from Bio.PDB import PDBParser
import numpy as np

# Parse a local PDB file path -- ensure the file exists at the specified path
parser = PDBParser(QUIET=True)

try:
    structure = parser.get_structure("structure", "5wb7.pdb")
except FileNotFoundError:
    print("Error: The PDB file was not found. Please verify the file name.")
    raise
except ImportError:
    print("Error: BioPython is not installed. pip install biopython")
    raise

# Access the first model and chain
models = list(structure)
if not models:
    raise ValueError("No models found in the PDB structure.")
model = models[0]

chains = list(model.get_chains())
if not chains:
    raise ValueError("No chains found in the model.")
chain = chains[0]

residues = list(chain.get_residues())
if not residues:
    raise ValueError("No residues found in the chain.")
```

## Common Misuse and Error Prevention

Note: This skill is for PDB files only. If the user provides an audio
recording (e.g., .mp3), it cannot be processed here.
\end{verbatim}
\end{small}

\subsubsection{Base Inference Editor: Edited \texttt{pdb} Skill}

The untrained base Qwen3-4B model replaces the entire skill with the following. No executable code remains. The ``Key Requirements'' section mistakenly names the GAIA task UUID \texttt{7dd30055-0198-452e-8c25-f73dbe27dcb8} as an invalid PDB identifier, and the ``Common Misuse'' section is wholly devoted to handling a calculus homework audio recording (\texttt{Homework.mp3}) that is unrelated to PDB processing.

\begin{small}
\begin{verbatim}
name: pdb
description: [unchanged from original]

# PDB (Protein Data Bank) File Processing Guide

## Key Requirements and Best Practices

- Always ensure the PDB ID is valid and accessible via the RCSB PDB database.
- Use the correct PDB ID (e.g., "5wb7", not
  "7dd30055-0198-452e-8c25-f73dbe27dcb8") -- the latter is a UUID or
  invalid identifier.
- This skill does NOT handle audio files. If a user provides an audio file
  (e.g., Homework.mp3) and asks for page numbers, this skill is NOT applicable.

## Common Misuse and Clarifications

- Do not use this skill to extract page numbers from audio recordings.
- Do not attempt to parse audio files as if they were PDB files.
- If a user asks about study materials from a lecture, state that the skill
  cannot process audio content.
- Always verify the nature of the input: if it is .mp3 or .wav, respond with:
  "I cannot extract page numbers from audio files."

## Example of Incorrect Use

User: "I have a file called Homework.mp3 with my professor reading the
       recommended pages. Can you tell me the page numbers?"
Correct response: "I cannot extract page numbers from audio files. This skill
  is designed for processing PDB files, not audio recordings."

[--- No code examples remain in the edited skill ---]
\end{verbatim}
\end{small}

The contrast is stark. The GRPO-trained editor adds \emph{code} that handles real failure modes---a missing file, a missing library---while preserving the entire original reference. The base inference editor removes all code and fills the skill with content copied from two specific benchmark questions, leaving it useless for any unseen PDB task.

\end{document}